\documentclass{article}
\usepackage{ijcai24}
\usepackage{todonotes}
\usepackage{times}
\usepackage{soul}
\usepackage{url}
\usepackage[hidelinks]{hyperref}
\usepackage[utf8]{inputenc}
\usepackage[small]{caption}
\usepackage{graphicx}
\usepackage{amsmath}
\usepackage{amsthm}
\usepackage{booktabs}
\usepackage{algorithm}
\usepackage{algorithmic}
\usepackage[switch]{lineno}

\usepackage{amssymb}
\usepackage{makecell}
\usepackage{multirow}

\newcommand\Rt[1]{\textcolor{black}{#1}}

\title{DMAT: A Dynamic Mask-Aware Transformer for Human De-occlusion}

\author{
    Anonymous Submission
}

\author{
Guoqiang Liang
\and
Jiahao Hu\thanks{Corresponding Author}
\and
Qingyue Wang\And
Shizhou Zhang
\affiliations
Northwestern Polytechnical University \\
\emails
gqliang@nwpu.edu.cn,
noreasonhhh@gmail.com,
li25722643@gmail.com,
szzhang@nwpu.edu.cn,
}

\begin{document}

\maketitle

\begin{abstract}
Human de-occlusion, which aims to infer the appearance of invisible human parts from an occluded image, has great value in many human-related tasks, such as person re-id, and intention inference. To address this task, this paper proposes a dynamic mask-aware transformer (DMAT), which dynamically augments information from human regions and weakens that from occlusion. First, to enhance token representation, we design an expanded convolution head with enlarged kernels, which captures more local valid context and mitigates the influence of surrounding occlusion. To concentrate on the visible human parts, we propose a novel dynamic multi-head human-mask guided attention mechanism through integrating multiple masks, which can prevent the de-occluded regions from assimilating to the background. Besides, a region upsampling strategy is utilized to alleviate the impact of occlusion on interpolated images. During model learning, an amodal loss is developed to further emphasize the recovery effect of human regions, which also refines the model's convergence. Extensive experiments on the AHP dataset demonstrate its superior performance compared to recent state-of-the-art methods. 


\end{abstract}

\section{Introduction}  

Human is practised in perceiving the whole object even when it is partially occluded \cite{kanizsa1979organization}. This ability helps us to understand visual scene input consisting of fragmented, incomplete and disordered objects, which is beneficial for many tasks, like intention inference. Therefore, image amodal completion, which aims to equip computers with this ability, has attracted lots of interest \cite{amodal_survey}. This paper focuses on amodal completion for human. In other words, we aim to recover the appearance content of invisible human parts given an occluded human image, also called \textbf{human de-occlusion} \cite{HDO}. Compared with other objects, especially rigid objects, human de-occlusion is much more complex because of the large variation of human pose and clothes. Therefore, modelling the visible context with arbitrary shapes and appearance becomes a key problem. 

Currently, sort-of-the-art (SOTA) human de-occlusion approaches are based on CNN architecture \cite{HDO,POISE}. For example, \cite{HDO} designed a stacked network and a parsing guided attention (PGA) module for mask completion and content recovery respectively. Although they obtain good performance, there exist some limitations. Limited by the receptive field of CNNs, global information is only gradually propagated across numerous layers. The resulted de-occluded regions show similar texture with neighbour regions, which extremely impairs human de-occlusion performance. Moreover, during inference, the elements between adjacent layers are connected via fixed weights rather than input-dependent weights. Although \cite{HDO} employed a PGA module to transmit human structure, it highly depends on inferred parsing modal masks and amodal masks, which refer to the visible and integrated portions of human, respectively. Since these parsing masks may contain additional background or occlusion information, they will deliver unreliable long-distance messages.

On the other hand, image inpainting, which aims to fill man-made holes with plausible contents, shares similarity with this task. \textit{Notably, image amodal completion aims to recover the appearance of occluded object parts while image inpainting is to fill holes and generate a reasonable image.} To exploit visible information sufficiently, many transformer-based methods have been proposed \cite{TFill,MAT,FcF}. Since this task usually utilizes all visible image regions to fill holes, straightly applying these models to human de-occlusion will lead to an \textit{attention-shifted} problem. In other words, the recovered content resembles the background more than the intended human subjects as shown in Figure \ref{fig:instance}.


\begin{figure*}
    \centering
    \includegraphics*[scale=0.45]{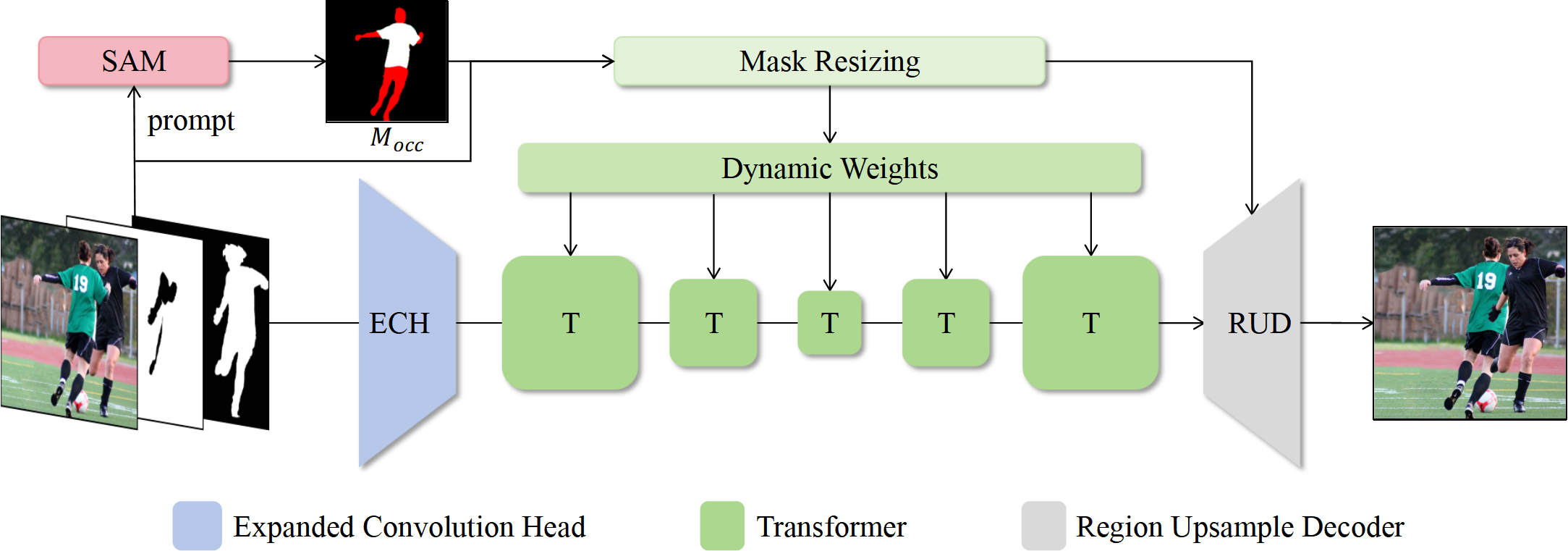}
    \caption{The proposed dynamic mask-aware transformer (DMAT) for human de-occlusion, which consists of an Expanded Convolution Head, a Transformer Body and a Region Upsampling Decoder. The visible mask and the amodal mask are delivered to the mask resizing module to guide the Transformer Body and the Region Upsampling Decoder.}
    \label{fig:framework}
\end{figure*}

To address these issues, we propose a \textbf{Dynamic Mask-Aware Transformer} (DMAT) to dynamically adjust the attention upon the input during modelling global and local context. Note that we focus on content recovery for simplicity. During training, we use the DMAT as a generator in a GAN framework. As shown in Figure \ref{fig:framework}, our DMAT consists of a head, a Transformer body and a decoder, which are for token embedding, global context modelling and resolution restoring. Instead of using deeply stacked convolutional layers, we employ the transformer due to its ability for capturing long-range context. Considering the computational intensity, we opt for the Swin-Transformer \cite{liu2021swin} to facilitate efficient training. However, it gradually acquires global context through shifted window partitioning, leading to potential shortcomings in capturing comprehensive global information. To address this limitation, we design a \textit{Expanded Convolution Head} (ECH), which enlarges the kernel size of partial convolutions. Thus, the resulted tokens encapsulate a more comprehensive representation of visible information, which mitigates the impact of occlusion close to the invisible human regions under the CNN local computation paradigm. Concurrently, the tokens with sufficient local context can bolster global context aggregation, particularly in shifted window partitioning. 
%

To mitigate the attention-shifted problem, we design \textit{Dynamic Human-Mask Guided Attention} mechanism (DHMGA) in the transformer body, which aims to refocus on visible human parts rather than background or occluded regions when modelling global context. To achieve this goal, the DHMGA involves three masks, i.e. human modal mask, invisible mask and occlusion mask. Through their combination, the whole image is split into four parts, the visible human parts, the human de-occluded regions, the occluding objects and other areas, each of which is assigned different weight. Therefore, the DHMGA can enhance beneficial information propagation, especially from visible human parts.

In human de-occlusion, occluding objects are closely around human subjects. When using traditional decoder, they will contaminate the recovered human parts, which hurts the de-occlusion performance.
To alleviate this adverse influence, we introduce an innovative \textit{Region Upsampling} (RU) strategy for the decoder, which meticulously partitions and upsamples the whole image as two distinct regions, i.e. human regions and others. Besides, we introduce an \textit{amodal loss} by adding the human amodal mask, which confine the recovery region exclusively to the human rather than others.

To validate the proposed method, we have conducted extensive experiments on the Amodal Human Perception (AHP) dataset \cite{HDO}, whose results show the effectiveness of DMAT. 

We highlight our main contributions as follows: 
\begin{itemize}
\item To tackle the attention-shifted problem, we propose a novel dynamic human-mask guided attention, emphasising human regions and ignoring occluding objects. 
\item We build a DMAT framework by integrating the DHMGA augmented transformer body with ECH, which incorporates more visible local information to complement global context modelling.
\item Extensive experiments on the AHP dataset demonstrate that the proposed model achieves a new SOTA performance for human de-occlusion.
\end{itemize} 

\section{Related Work}

\subsection{Image Amodal Completion}
Generally, image amodal completion consists of amodal segmentation, amodal appearance completion and order perception. Amodal segmentation \cite{amodal_seg,SAS} estimates the full shape of each partially occluded object, which is usually represented as a binary segmentation mask, including its invisible area. Through a CNN-based network, APSNet \cite{amodal_seg} produced a pixel-level semantic segmentation map and an instance-level amodal segmentation mask simultaneously. Amodal appearance completion infers the plausible appearance of invisible portion from visible object parts. In \cite{monnier2021unsupervised}, an unsupervised learning framework was presented to jointly learn the object prototypes and occlusion/transformation predictors to explain images. In order perception, both occlusion order and depth order are produced. For example, \cite{orders} introduced a geometric order prediction network and a dense depth prediction network. In recent works \cite{amodal_seg}, these problems are studied jointly.


 
\paragraph{Human De-occlusion.} Human de-occlusion \cite{HDO} is a category-specific amodal appearance completion task. In other words, it needs to infer both the invisible human regions and their reasonable appearance content. This task is much more difficult due to the large variation of human pose and clothes compared to other objects. To extract a human amodal mask, POISE \cite{POISE} integrated noisy human modal mask from a segmentation model and human joint estimation. Moreover, a two-stage framework is developed to jointly complete the human amodal segmentation and content recovery \cite{HDO}. Limited by the receptive field of CNN, it is tough to model global context sufficiently. Compared with these, we focus on content recovery and combine the CNN and transformer to fully utilize the global context.

\subsection{Image Inpainting}
\paragraph{CNN-based Inpainting.} 
\cite{pathak2016context} first introduced GAN \cite{GAN} into image inpainting. To reduce color discrepancy and blurriness, partial convolution (PConv) \cite{PConv} and gated convolution \cite{GatedConv} were proposed to conduct convolution only on valid pixels. However, their performance is limited by the small receptive field. To address this problem, the Fast Fourier Convolution (FFC) was applied to aggregate global context \cite{lama}. Moreover, \cite{FcF} designed a new FFC structure while \cite{SHGAN} incorporated spectral information to generate high-quality images.

\paragraph{Transformer-based Inpainting.} 
Recently, due to the powerful ability of capturing long-range dependence, transformer \cite{ViT,liu2021swin} has gained popularity in image inpainting. For example, TFill \cite{TFill} employed a restrictive CNN for weighted token representation. Due to the large number of image tokens, TFill based on original transformer is computationally expensive. Based on the more efficient Swin-Transformer, a dynamic mask is designed to aggregate global context only from partial valid tokens \cite{MAT}. Besides, diffusion models \cite{DDRM,RePaint} have been applied in image inpainting. Note that none of them have been applied for human de-occlusion.

Compared with the above methods, we address the attention-shifted problem in human de-occlusion through a novel CNN-Transformer framework. Our approach incorporates a specialized convolution head and a dynamic human-mask guided attention mechanism, which dynamically adjusts the attention according to the various importance of image tokens during modelling global context.



\section{Method}
This section first gives the overall structure. Then, we describe the three main proposed modules. Finally, we detail the reivsed amodal loss.

\subsection{Overview}\label{sec:overview}
As shown in Figure \ref{fig:framework}, our proposed DMAT consists of an expanded convolution head (ECH), a transformer body augmented by dynamic human-mask guided attention (DHMGA) and a region upsampling decoder (RU), which thoroughly combines the advantages of convolution and transformer. In particular, the ECH aims to extract tokens with rich local context, which helps the transformer to aggregate global context and reduce the influence of occlusion. Through our proposed DHMGA, the transformer body can dynamically focus on the visible human body when modelling long-range relationships. Finally, an RU strategy is designed to protect the de-occluded human region when increasing the spatial size.

Like the content recovery model in \cite{HDO}, our input is a five-channel tensor $x'=stack(x \odot M_{vis}, M_{vis}, M_{amodal})$. $x$ represents an occluded human image while $M_{vis}$, $M_{amodal}$ represent its visible mask and amodal mask respectively. These two masks can be obtained by the mask completion network in HDO. Here, we directly utilize the ground-truth masks to focus on content recovery. Besides, we also employ the SAM \cite{SAM} to get $M_{occ}$ by randomly selecting 5 points in $M_{inv}=1-M_{vis}$ as position prompts. 

\begin{figure}
    \centering
    \includegraphics*[scale=0.32]{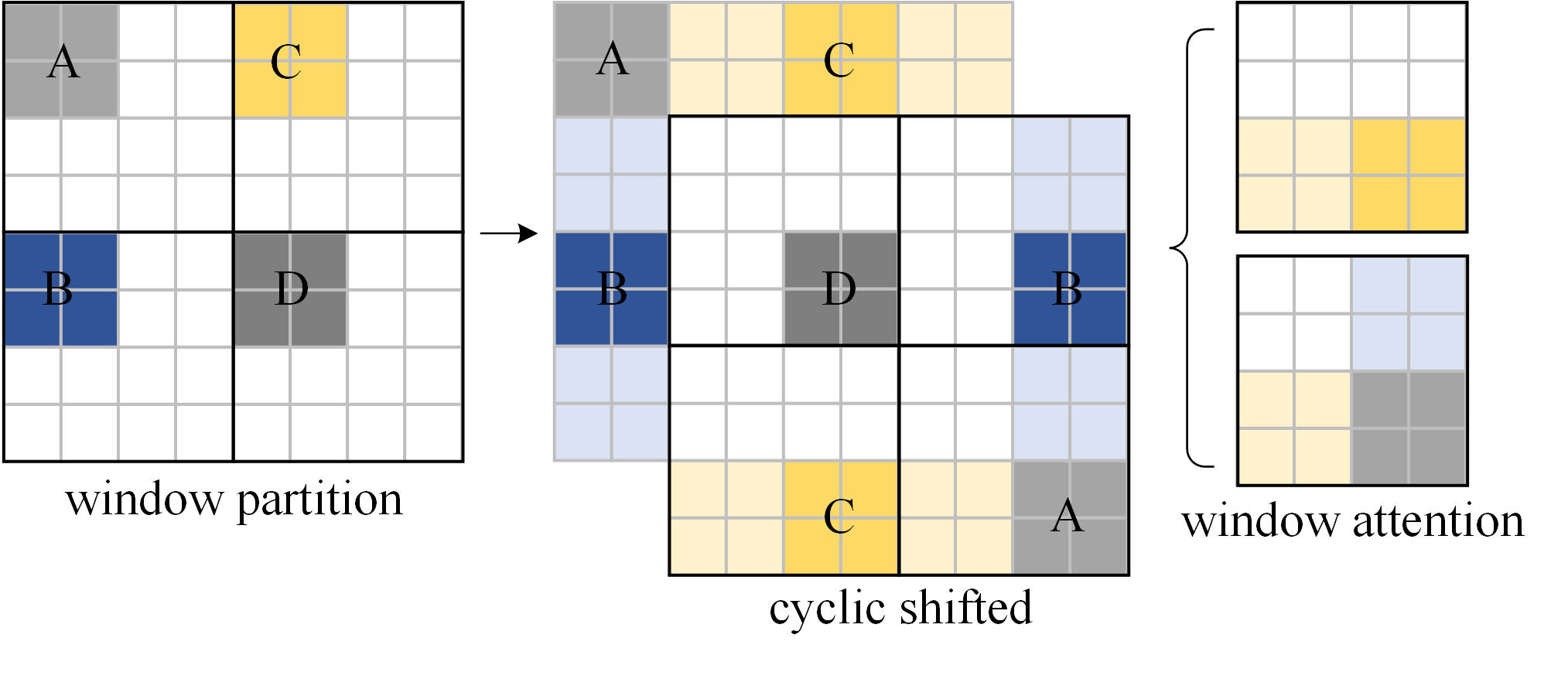}
    \caption{
    Illustration of self-attention in shifted window partitioning. 'A, B, C, D' represent different regions. 
    }
    \label{fig:windows}
\end{figure}

\subsection{Expanded Convolution Head}
To reduce computational cost, we utilize the Swin-Transformer \cite{liu2021swin} to capture global context. The Swin-transformer uses shifted window partitioning to establish cross-window connections and acquires global context progressively. However, the window-based self-attention module brings inherent limitations for aggregating global context. As illustrated in Figure \ref{fig:windows}, block $A$ does not engage in attention processing for blocks $B, C, D$. On the other hand, when using traditional convolution for token embedding, occluding objects may also be embedded in visible tokens since they are always around human. In other words, the appearance of occluding objects will implicitly be used for de-occlusion, which hurts the de-occlusion performance. 


\begin{figure}
    \centering
    \includegraphics*[scale=0.15]{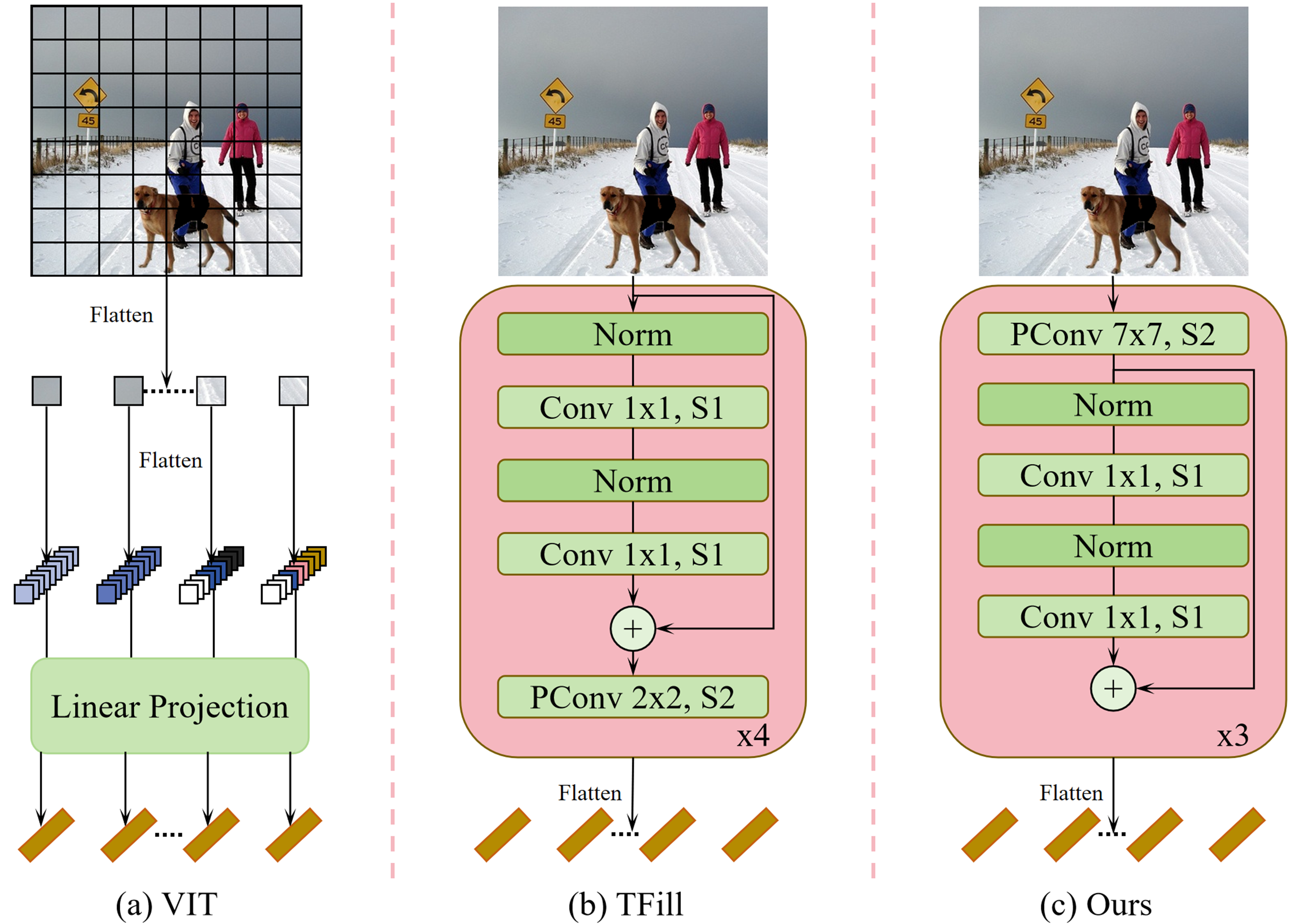}
    \caption{\textbf{Token representation.} (a) Patch to tokens. (b) Restrictive Receptive Field (RF) feature to tokens. (c) Our Expanded RF feature to tokens. Our tokens have a large RF and use a stacked $(\times3)$ CNN embedding. These tokens with richer local features will promote global context modelling in the swin-transformer body.}
    \label{fig:encoder}
\end{figure}

To address the above challenges, we design an \textit{Expanded Convolution Head} (ECH) to incorporate more local valid context. Specifically, we enlarge the kernel size of partial convolution in the three convolution blocks for token embedding as shown in Figure \ref{fig:encoder}. Furthermore, we optimize the network architecture by positioning the down-sampling layer before residual connection, which reduces the computational complexity compared with TFill \cite{TFill}. This combination of large kernel size with partial convolution makes tokens enriched with substantial local visible context. 
\Rt{Additionally, through the strategic implementation of shifting window partitioning, tokens within the same window progressively contribute to the heightened local context, facilitating the aggregation of global context.} Meanwhile, window-based attention is less susceptible to cross-contamination.


\subsection{Dynamic Human-Mask Guided Attention} \label{sec:DHMGA}

Due to the existence of occlusion, the fidelity of tokens is different. To handle this problem, \cite{MAT} introduced multi-head contextual attention, which conducts interaction only using valid tokens. This attention works in image inpainting, which is just to generate reasonable images. However, in human de-occlusion, the recovered appearance should be similar to visible human regions instead of background. Thus, the attention, which can not discriminate human from background, is not suitable. Actually, directing applying this to human de-occlusion leads to an attention-shifted problem, where the color of the de-occluded region is akin to the surrounding background, as shown in Figure \ref{fig:instance}.



To address the above problems, we design a novel attention mechanism termed \textit{Dynamic Human-Mask Guided Attention} (DHMGA), which pays more attention to valid human regions and less to occlusion. In detail, the DHMGA involves three masks $M_{inv}, M_{modal}$ and $M_{occ}$, which provide coarse information for rightly dividing a whole image. For a single head attention, the DHMGA can be formulated as 
\begin{equation}\label{eq:DHMGA}
    \begin{aligned}
        Attn(Q,K,V) & = Softmax(\frac{QK^T+\sum_{t}\alpha_{t}\beta_t}{\sqrt{d_k}})V, 
    \end{aligned}    
\end{equation}
where $Q, K, V$ are the query, key, value matrices, $\frac{1}{\sqrt{d_k}}$ is the scaling factor. $t \in \{inv, modal, occ\}$ denotes the mask type. $\beta_{t} \in R^{wsz^2}$ is prior weight depending on each input mask, where $wsz$ denotes the window size. The $\beta_t$ is calculated as:
\begin{equation}
    \beta_{t,n} = 
    \begin{cases}
           0 & \text{if } M_{t,n} = 0 , \\
        \tau_t & \text{if } M_{t,n} = 1 , 
    \end{cases}
\end{equation}
where $\tau_t$ is an integer representing the various importance of tokens from different regions. According to our experiments, we set them to -100.0, 30.0 and -100.0 for $M_{inv}, M_{modal}, M_{occ}$ respectively. 
To find a treat of these regions' weights, we additionally introduce learnable parameters $\alpha_t \in R^{wsz^2}$ for these masks, which facilitates the dynamic adjustment of weights for these tokens. 



\paragraph{Updating Strategy for Masks.} Following the released implementation of MAT \cite{MAT}, we adopt the same way for updating $M_{inv}$ in the transformer downsampling stage while directly using the original invisible mask as $M_{inv}$ in the upsampling stage. Other masks are not updated because they need to maintain modal and occluded regions, which will be emphasised or weakened in attention.

\begin{figure}
    \setlength{\belowcaptionskip}{-0.cm}
    \centering
    \includegraphics*[scale=0.35]{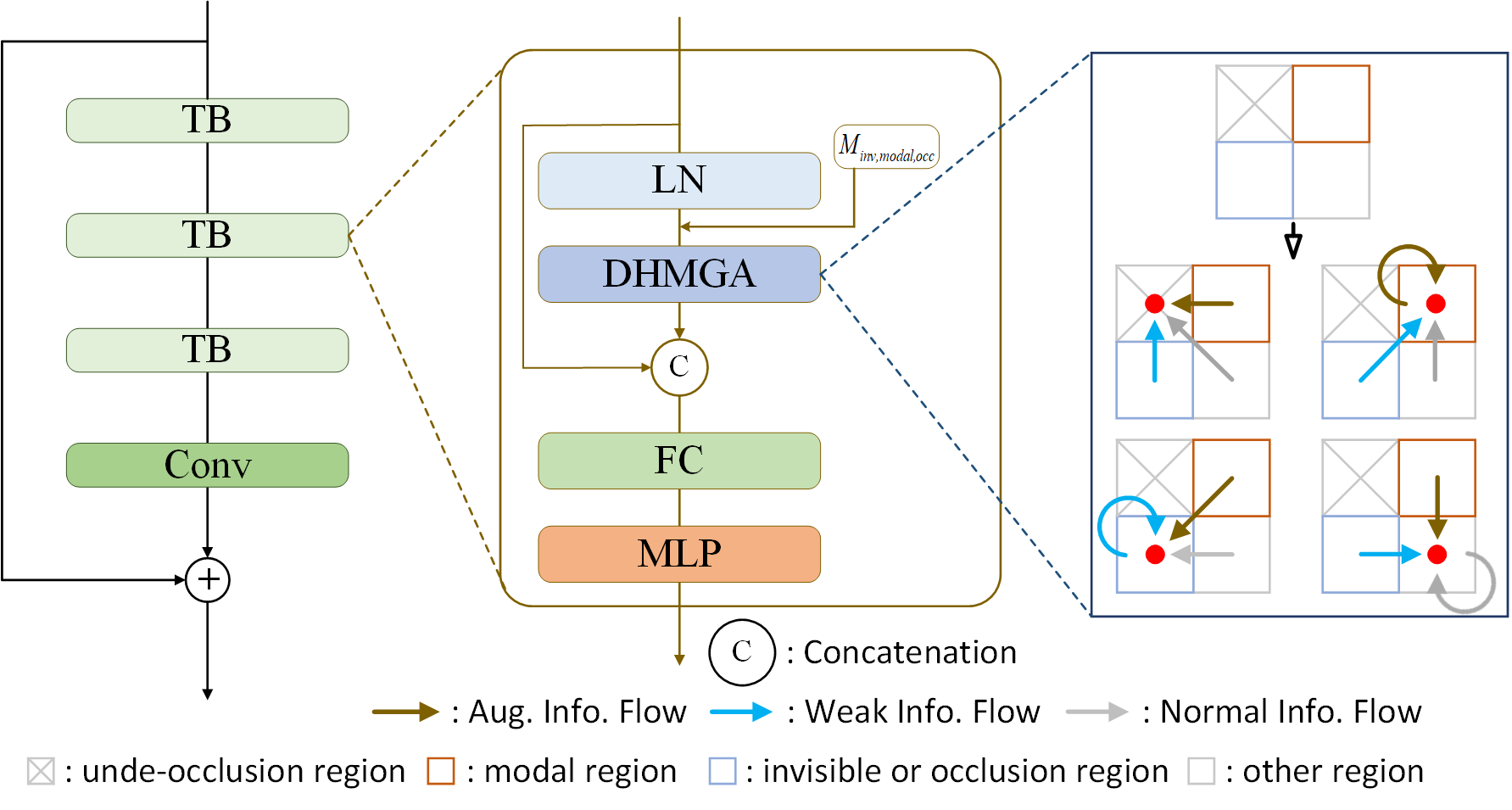}
    \caption{\textbf{Structure of a single transformer stage.} 
    "TB" refers to a transformer block, whose core module is the proposed DHMGA, which aggregates tokens from different regions by different weights. See text for more details}
    \label{fig:DHMGA}
\end{figure}

According to Eq. \eqref{eq:DHMGA}, the composition of these three masks will classify a whole image into four kinds, which are assigned different weights. Therefore, various attention is paid to these regions when propagating global context. An illustrative example is shown in Figure \ref{fig:DHMGA}, where red-brown arrows represent the augmented flow from human modal regions, the azure arrows denote the weakened flow from invisible or occluded regions, and gray arrows are normal flow from other region tokens. Moreover, the aggregation weights for un-de-occluded tokens are nearly 0. In a word, our DHMGA can augment the information from visible human regions and weaken that from invisible regions and objects covering human, which finally addresses the attention-shifted problem.

\subsection{Region Upsampling Decoder}
Like previous works \cite{yi2020contextual,TFill}, a gradual upsampling decoder is employed to generate high-resolution de-occlusion images. Generally, these upsampling blocks consist of a bilinear interpolation followed by several convolutional layers. However, this original interpolation leads to degraded images due to the inherent occlusion around the de-occluded regions. Moreover, compared to commonplace objects, the contextual dependencies between human and background are relatively weak, which is less useful for appearance inference.

In light of these considerations, we propose a novel \textit{Region Upsampling} (RU) strategy to protect recovered humans,
\begin{align}
    \hat{x} = & Up(\hat{x} \odot M) \odot Up(M) + \notag \\
        & Up(\hat{x} \odot (1- M)) \odot Up(1 - M)
\end{align}
where $M$ represent the amodal mask $M_{amodal}$. By upsampling the human region and background individually, we can get a better effect for boundaries and recovered human regions.  

\begin{table*}[t]
    \setlength{\abovecaptionskip}{0.1cm}
    \begin{center}
    \begin{tabular}{l|rrrrrr|rrrrr|rr}
        \toprule
    \multirow{2}{*}{} & \multicolumn{6}{c|}{Syn. FID$\downarrow$} & \multicolumn{5}{c|}{Real FID$\downarrow$} & \multirow{2}{*}{\makecell[r]{Syn. \\ HFID}} & \multirow{2}{*}{\makecell[r]{Real \\ HFID}}
    \\ \cline{2-7} \cline{8-12}
    \makebox[0.04\textwidth][l]{Methods}  & 
    \makebox[0.04\textwidth][r]{Total}   & \makebox[0.04\textwidth][r]{0-10\%}  &
    \makebox[0.04\textwidth][r]{10-20\%} & \makebox[0.04\textwidth][r]{20-30\%} & 
    \makebox[0.04\textwidth][r]{30-40\%} & \makebox[0.04\textwidth][r]{40-50\%} &
    \makebox[0.04\textwidth][r]{Total}   & \makebox[0.04\textwidth][r]{0-10\%}  & 
    \makebox[0.04\textwidth][r]{10-20\%} & \makebox[0.04\textwidth][r]{20-30\%} & 
    \makebox[0.04\textwidth][r]{30-40\%} 
    & \makebox[0.04\textwidth][r]{} & \makebox[0.04\textwidth][r]{}
    \\ \midrule
    PConv       & 15.34   & 8.91    & 22.61    & 33.47    & 45.40    & 79.35   
                & 19.72   & 14.06   & 22.46    & 23.35    & 54.68    & 20.56 & 44.42
                \\
    DeepFillv2  & 16.60   & 10.99   & 23.81    & 36.46    & 48.65    & 86.21  
                & 15.92   & 14.43   & 19.27    & 16.25    & 47.26    & -     & -
                \\
    DSNet       & 13.23   & 8.17    & 19.57    & 29.00    & 40.76    & \underline{62.59}  
                & 14.02   & 10.30   & 17.18    & 17.73    & 38.67    & 20.28 & 59.23
                \\
    HDO         & 13.85   & -       & -        & -        & -        & -       
                & 19.49   & -       & -        & -        & -        & -     & -
                \\
    RePaint     & 17.19   & 13.58   & 26.84    & 40.01    & 50.14    & 83.28  
                & -       & -       & -        & -        & -        & -     & -
                \\
    MAT         & 15.08   & 9.93    & 22.54    & 32.27    & 43.90    & 75.55  
                & 16.72   & 14.18   & 21.83    & 17.25    & 42.51    & 19.52 & 47.85
                \\ 
    TFill       & \underline{12.93}   & \underline{7.68}    & \underline{17.90}    & \underline{27.49}    & \underline{39.15}    & 70.91  
                & \underline{11.65}   & 9.06    & 14.59    & \underline{11.20}    & \underline{34.37}    & \underline{19.08} & \underline{41.59}
                \\ 
    FcF         & 17.42   & 13.27   & 25.92    & 39.03    & 54.90    & 91.04  
                & 12.55   & \underline{8.53}    & 16.41    & 12.79    & 45.71    & -     & -
                \\ 
    MIGAN       & 16.03   & 10.35   & 24.37    & 35.71    & 49.53    & 88.96  
                & 11.73   & 9.50    & \underline{12.49}    & 13.87    & 48.61    & 23.09 & 42.85
                \\ \midrule[0.25pt]
    DMAT(base)     & 11.83   & 7.06    & 16.53    & 24.95    & 36.54    & 59.88  
                & 11.43   & \textbf{7.05}& 12.92  & 10.04    & 30.13 & 17.51 & 42.62 
                \\
    DMAT        & \textbf{10.99}   & \textbf{6.59}    & \textbf{14.73}    & \textbf{23.64}    & \textbf{33.23}    & \textbf{57.04}  
                & \textbf{8.75}    & 7.57    & \textbf{9.72}    & \textbf{8.90}     & \textbf{28.80}   & \textbf{16.34} & \textbf{39.43} 
                \\             
                \bottomrule
        \end{tabular}
        \caption{
        Performance comparisons on the AHP dataset with ground-truth amodal and modal masks. 'Syn.' and 'Real' denote the synthesized and real validation set. `$0-10\%$' represents the occlusion rate while `total' denotes all images.
        }
        \label{tab:ratio_AHP}
    \end{center}
\end{table*}

\subsection{Amodal Loss}
Since our DMAT is based on a GAN, we adopt the non-saturating adversarial loss. We also employ the popular reconstruction loss, perceptual loss, and style loss for easier optimisation. In our initial experiments, their combination impedes the model's convergence state because they are originally for recovering whole images. Considering that we concentrate on human appearance recovery, we add the human amodal mask to these loss functions, which confines the recovery region exclusively to human. We term this resulted loss as \textit{amodal loss}.
\paragraph{Adversarial Loss.} We calculate the adversarial loss as
\begin{align}
    L_G = -& E_{\hat{x}}[log(D(\hat{x} \odot M))], \\
    L_D = -& E_x[log(D(x \odot M))] - \notag \\ 
           & E_{\hat{x}}[1 - log(D(\hat{x} \odot M))],
\end{align}
where $x$ and $\hat{x}$ are real and de-occluded image. $D$ represents the discriminator, and $M$ is the human amodal mask.

\paragraph{Reconstruction Loss.} The reconstruction loss aiming at pixel-level approximation is calculated as 
\begin{equation}
    L_1 = \sum_{i,j\in M}{\Vert \hat{x}_{ij} - x_{ij} \Vert_1},
\end{equation}
where $i, j$ denotes a pixel location in human amodal mask.

\paragraph{Perceptual Loss.} The perceptual loss is formulated as 
\begin{equation}
    L_P = \Vert \phi_5(\hat{x} \odot M) - \phi_5(x \odot M) \Vert_1,
\end{equation}
where $\phi_5(\cdot)$ denotes the activation of layer $conv_{5}$ in a pre-trained VGG-16 network \cite{VGG}. 

\paragraph{Style Loss.} Following \cite{HDO,PConv}, we calculate the style loss as 
\begin{align}
    L_{S} = \sum_{i} \Vert & Gr(\phi_i(\hat{x} \odot M)) - Gr(\phi_i(x  \odot M)) \Vert_1,   
\end{align}
where $Gr$ denotes the auto-correlation operation.

In a word, the overall amodal loss can be formulated as:
\begin{equation}
    L = \lambda_1 L_1 + \lambda_2 L_G + \lambda_3 L_D + \lambda_4 L_P + \lambda_5 L_{S},
\end{equation}
where $\lambda_1=15$, $\lambda_2=0.06$, $\lambda_3=0.6$, $\lambda_4=1$ and $\lambda_5=150$ are coefficients chosen by experiments. 



\section{Experiments}

\begin{figure*}[!ht] 
    \setlength{\abovecaptionskip}{-0.2cm}
    \setlength{\belowcaptionskip}{-0.5cm}
    \centering
    \includegraphics*[scale=0.5]{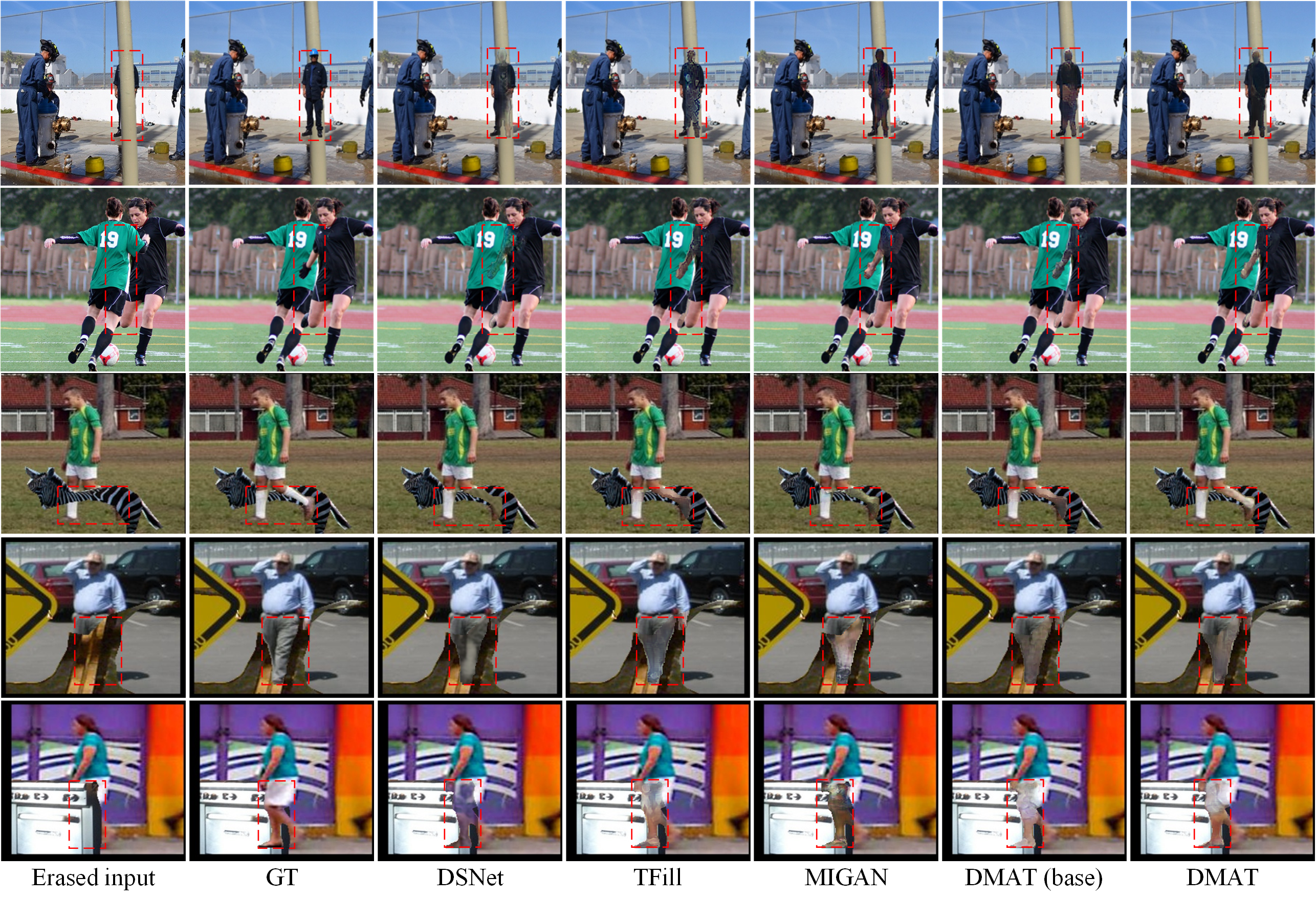}
    \caption{
    Comparison of visual examples. DMAT generates more reasonable appearances and human structures. Please zoom in to see details.
    }
    \label{fig:instance}
\end{figure*}

\subsection{Dataset and Evaluation Metrics}
\paragraph{AHP Dataset.} To validate the proposed DMAT, we employ the AHP dataset \cite{HDO}, whose distribution of human occlusion ratios can be controlled manually. Following the original way of selecting instances and occlusion ratio, we generate 53, 336 occluded human images for training. To reduce computation, this generated training set is fixed rather than synthesizing new occluded human images repeatedly during each training batch in the original paper. For testing, the AHP consists of a synthesized validation set and a real validation set. The former contains artificially synthesized 891 images while the latter includes 56 real images from real scenes. 

\paragraph{Evaluation Metrics.}
Like previous methods, we adopt the $l_1$ distance and Fr\'{e}chet Inception Distance (FID) \cite{FID}, which measures the similarity between the whole ground-truth image and the de-occluded result. Since this task focuses on human, we also report the Human FID (HFID) to only measure the similarity of human regions.

\subsection{Implementation Details}
For ECH, we set the numbers of convolution channels to $\lbrace64, 128, 256\rbrace$. For each transformer body, both the number of convolution channels and the dimension of a fully connected layer are set to 128. We employ $\lbrace2, 2, 6, 2, 2\rbrace$ blocks for 5-level transformer groups, whose window size is $\lbrace8, 8, 4, 8, 8\rbrace$ respectively. Following \cite{TFill,zheng2021tfill}, a region upsampling decoder with channels $\lbrace256, 128, 6\rbrace$ is used to de-occlude human images. The Patch-GAN \cite{PatchGAN} with four convolution layers is adopted as our discriminator. We use the Adam optimizer, whose learning rate is set to $\lbrack 1 \times 10 ^ {-2}, 1 \times 10 ^ {-3}, 2 \times 10 ^ {-4}, 1 \times 10 ^ {-4}, 5 \times 10 ^ {-5}\rbrack$ in the iterations [1k, 20k, 60k, 150k] respectively. We use the Exponential Moving Average \cite{EMA} with a rate $1\times 10^{-5}$ for learning rate decay. The maximum iteration number is 200k and the batch size is 16.

As stated in section \ref{sec:overview}, our input consists of occluded image $x$, visible mask $M_{vis}$, and amodal mask $M_{amodal}$. We derive the two masks from genuine images and artificially selected occlusions. Besides, we use the SAM \cite{SAM} to get $M_{occ}$ by randomly selecting 5 points in $M_{vis} = 1 - M_{inv}$ as position prompt. All experiments are carried out on a server with an NVIDIA RTX 3090 GPU. 


\subsection{Quantitative Results}
First, we compare the proposed DMAT with some SOTA methods, including PConv \cite{PConv}, DeepFillv2\cite{GatedConv}, DSNet \cite{wang2021dynamic}, HDO \cite{HDO}, RePaint \cite{RePaint}, MAT \cite{MAT}, TFill \cite{TFill}, FcF \cite{FcF} and MIGAN \cite{sargsyan2023mi}. The HDO is specially designed and achieves the best performance for human de-occlusion, while others are originally for image inpainting. Some of these methods, like TFill, employ a two-stage network, where one infers the content and the other is to refine appearance to generate high-quality images. Since we focus on the first stage, we only compare DMAT with the first-stage networks in these two-stage methods. Note that it is very straightforward to combine our work with the second network. Besides, since the lack of norm layers in MAT causes poor performance, we re-add the $InstanceNorm$ in our experiments. We use the released code to run our experiments and report the results.

The results of all the above methods on AHP are illustrated in Table \ref{tab:ratio_AHP}, where the `Syn.' and `Real' denote synthesized validation and real validation set respectively. `$0-10\%$' represents the occlusion rate while `total' denotes all images. `DMAT(base)' refers to a DMAT baseline without any improved module. From this table, we can see that DMAT outperforms all other methods. In particular, compared with the second-best results, the margin of DMAT is over 1.93 and 2.9 on synthesized and real validation sets respectively in total. Moreover, the margin becomes larger with the increase of occlusion ratio. This shows our method can still understand the whole human even with heavy occlusion. The improvement of Real FID is similar to the Syn. FID, which illustrates the generalization ability of DMAT. Furthermore, we 
compare DMAT with some methods in terms of HFID, whose results are shown in the last two columns of Table \ref{tab:ratio_AHP}. Like FID, our DMAT obtains the lowest scores, which validate the recovery effect of human. 

\subsection{Qualitative Results}
Figure \ref{fig:instance} gives some samples from DSNet, TFill, MIGAN, DMAT(base) and our DMAT, where GT denotes ground-truth. In these samples, although the three compared methods can produce plausible profiles, the color of recovered parts tends to be similar to the background, like the blue leg of DSNet in the second sample, the pant in the last sample, the leg with grass color in the third sample of MIGAN. In contrast, our DMAT gets the correct color even under heavy occlusion, \textit{e.g.} the white pants of the woman. In a word, our DMAT obtains superior visual results, even when significant semantic human information is missing.


\begin{table}[tb]
    \setlength{\abovecaptionskip}{0.1cm}
    \setlength{\belowcaptionskip}{-0.cm}
    \begin{center}
        \begin{tabular}{cccc|rr}
            \toprule
            \makebox[0.02\textwidth][c]{ECH} & \makebox[0.07\textwidth][c]{DHMGA} 
          & \makebox[0.02\textwidth][c]{RUD} & \makebox[0.02\textwidth][c]{AL}
          & \makebox[0.1\textwidth][r]{Syn. FID $\downarrow$}   
          & \makebox[0.1\textwidth][r]{Real FID $\downarrow$}    \\ \midrule
                           &                &              &              & 11.83    & 11.43     \\
            $\checkmark$   &                &              &              & 11.23    & 10.47     \\
                           &  $\checkmark$  &              &              & 11.18    & 11.04     \\ 
                           &                & $\checkmark$ &              & 11.53    & 10.00     \\
            $\checkmark$   &  $\checkmark$  &              &              & 11.16    & 9.91      \\             
            $\checkmark$   &                & $\checkmark$ &              & 11.18    & 10.04     \\
            $\checkmark$   &  $\checkmark$  & $\checkmark$ &              & 11.17    & 10.10     \\
            $\checkmark$   &  $\checkmark$  & $\checkmark$ & $\checkmark$ & \textbf{10.99}    & \textbf{8.75}   \\ \bottomrule
            \end{tabular}
            \caption{Effect of proposed modules, where AL is the amodal loss.}
            \label{tab:all}
    \end{center}
    \end{table}
\subsection{Ablation Study} \label{sec:Ablation Study}
Finally, we show the effect of the proposed modules and some settings for ECH and DHMGA. 
\paragraph{Effect of proposed Modules.} Our DMAT consists of three modules. To investigate their effectiveness, we have done an ablation study by adding each module to the redesigned human de-occlusion architecture, whose results are presented in Table \ref{tab:all}. From this table, we can find that adding each module individually can enhance the performance. However, directly combining them can not get obvious improvement, especially on the real validation set. We attribute this to the convergence issue since original loss functions aim to recover the whole image instead of human. As shown in the final row, the introduction of amodal loss leads to the best performance, which means that the model obtains a better convergence state.

\begin{table}[tb]
    \setlength{\abovecaptionskip}{0.1cm}
    \setlength{\belowcaptionskip}{-0.5cm}
    \begin{center}
        \begin{tabular}{l|rr|rr|r}
            \toprule
            \multirow{2}{*}{} & \multicolumn{2}{c|}{Syn.} & \multicolumn{2}{c|}{Real} 
            & \multirow{2}{*}{\makecell[r]{P(M)}} \\
            \cline{2-3} \cline{4-5}
            \makebox[0.05\textwidth][l]{Kernel}  
            & \makebox[0.03\textwidth][r]{$l_1\downarrow$} & \makebox[0.03\textwidth][r]{FID $\downarrow$} 
            & \makebox[0.03\textwidth][r]{$l_1\downarrow$} & \makebox[0.03\textwidth][r]{FID $\downarrow$} 
            & \makebox[0.03\textwidth][r]{}  \\ \midrule
        $\lbrack2, 2, 2\rbrack$ & 0.00751          & 11.55          & 0.00403          & 10.36 & 11.05 \\
        $\lbrack7, 7, 5\rbrack$ & 0.00738          & 11.59          & 0.00388          & 9.33 & 11.63 \\
        $\lbrack7, 7, 7\rbrack$ & 0.00767          & \textbf{10.99} & \textbf{0.00373} & \textbf{8.75} & 12.02 \\
        $\lbrack9, 7, 5\rbrack$ & 0.00736          & 11.26          & 0.00403 & 10.23  & 11.66 \\
        $\lbrack11, 7, 7\rbrack$& \textbf{0.00735} & 11.29          & 0.00412 & 10.31  & 12.09 \\
        \bottomrule
        \end{tabular}
        \caption{Influence of the kernel size of PConv layers in ECH. "Kernel" denotes the different sizes, where the three values correspond to the three layers. "P" is the training model parameter.}
        \label{tab:encoder}
    \end{center}
    \end{table}
\paragraph{Influence of Convolution Kernel Size.}
In ECH, we enlarge the kernel size of the three PConv layers. To investigate its influence, we have done experiments using various sizes, whose results are given in Table \ref{tab:encoder}. From it, we can find that increasing the kernel size can promote performance in the beginning. However, too large kernel may contain too much background, which reduces the FID scores. Therefore, we set the size of the three kernels to $\lbrack7, 7, 7\rbrack$, which achieves the best FID.



\begin{table}[tb]
    \setlength{\abovecaptionskip}{0.1cm}
    \setlength{\belowcaptionskip}{-0.cm}
    \begin{center}
        \begin{tabular}{ccc|rr}
            \toprule
            \makebox[0.04\textwidth][c]{modal} & \makebox[0.04\textwidth][c]{inv} 
            & \makebox[0.04\textwidth][c]{occ} & \makebox[0.1\textwidth][r]{Syn. FID $\downarrow$}   
                                               & \makebox[0.1\textwidth][r]{Real FID $\downarrow$}    \\ \midrule
                           &                  &                & 11.53    & 10.24 \\
            $\checkmark$   &                  &                & 11.18    & 9.06  \\
            $\checkmark$   &  $\checkmark$    &                & 11.61    & 9.91  \\ 
                           &                  & $\checkmark$   & 11.66    & 9.97 \\
            $\checkmark$   &                  & $\checkmark$   & 11.24    & 10.04 \\
            $\checkmark$   &  $\checkmark$    & $\checkmark$   & \textbf{10.99} & \textbf{8.75}  \\ \bottomrule
            \end{tabular}
            \caption{Ablation study of the mask in DHMGA. 
            The $modal$, $inv$ and $occ$ denote the modal mask, invisible mask and occlusion mask. }
            \label{tab:attention}
    \end{center}
\end{table}
\begin{figure}[tb]
    \setlength{\abovecaptionskip}{-0.2cm}
    \setlength{\belowcaptionskip}{-0.cm}
    \centering
    \includegraphics*[scale=0.45]{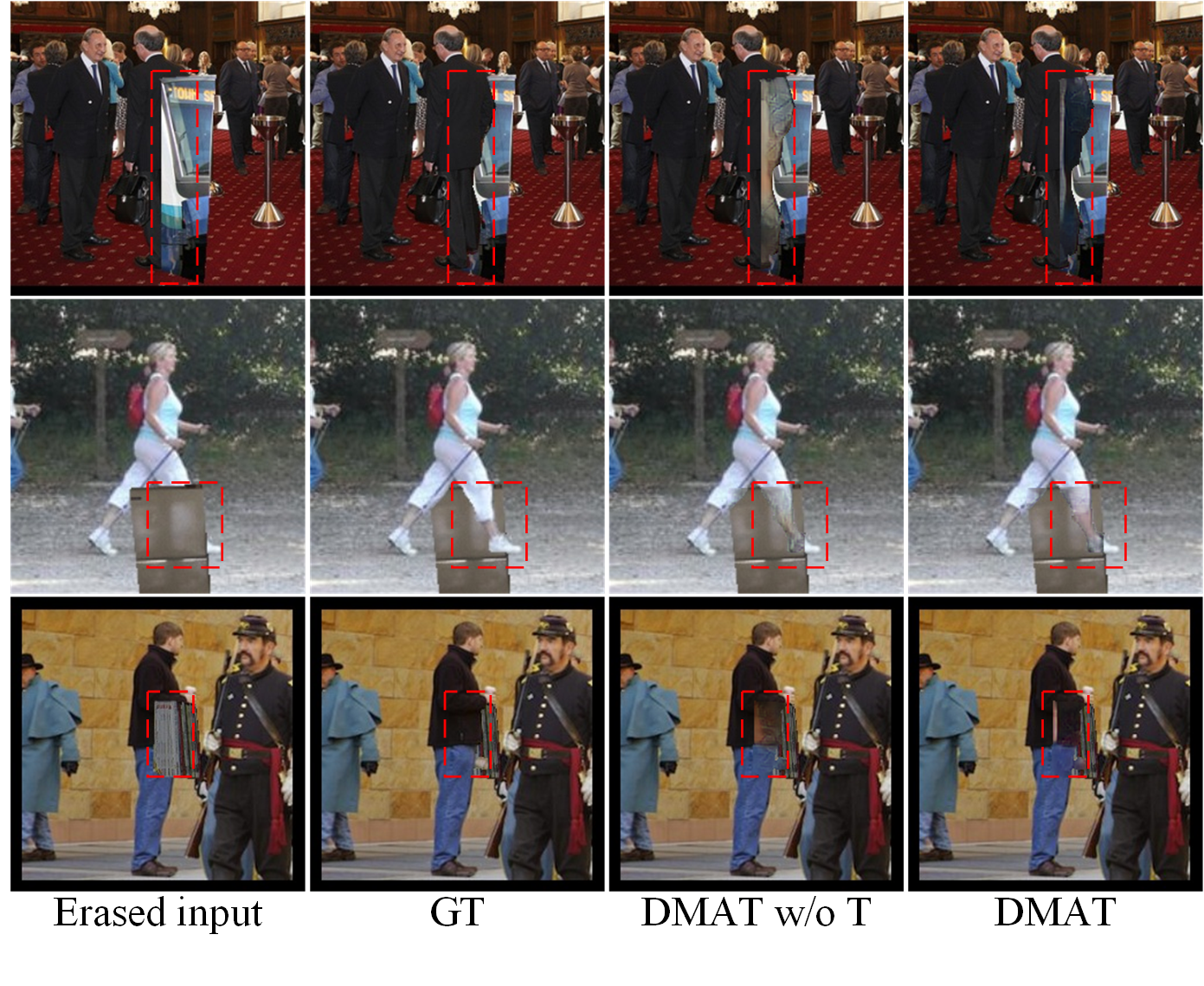}
    \caption{Influence of DHMGA. T represents the DHMGA module.}
    \label{fig:DHMGA_instance}
\end{figure}
    
\paragraph{Comparisons of Masks in DHMGA.}
The proposed DHMGA incorporates three masks to emphasize the human region. Table \ref{tab:attention} shows the FID results using different combinations of these masks. Since only using $M_{inv}$ raises the weight for occlusion token and damages performance, we do not show its result. In this table, using all masks obtains the best performance since it emphasizes the human modal token, and lessens the invisible and occlusion tokens simultaneously. Only using a portion of them even damages the performance. For example, when combining $M_{modal}$ and $M_{inv}$, the FID becomes worse since it also raises the occlusion token weight relatively. Besides, some examples are shown in Figure \ref{fig:DHMGA_instance}. After using the DHMGA, the recovered region becomes more similar to the ground-truth rather than the background or occlusion. This declares that the DHMGA makes the attention focus on human rather than background. In other words, the DHMGA solves the attention-shifted problem effectively.

More results are given in the supplementary materials. Please refer it for more visualization results.

\section{Conclusion}
Through a deep analysis of SOTA methods and challenges of human de-occlusion, we propose a novel Dynamic Mask-Aware Transformer network (DMAT). To propagate useful information from visible human regions, we develop three modules, i.e. the ECH to capture more local visible context, the DHMGA for correctly attending to human region and the RU to to mitigate occlusion effects. During training, we design amodal loss to enhance convergence. Extensive experiments on the AHP dataset show we achieved a new SOTA. In future, we will further investigate how to model symmetric relationship of human limbs. Besides, how to recover multiple persons at a time is also an interesting problem.


\bibliographystyle{named}
\bibliography{ijcai24}

\end{document}


\maketitle

\begin{table*}[hb]
    \begin{center}
    \begin{tabular}{l|rrrrrr|rrrrr}
        \toprule
    \multirow{2}{*}{} & \multicolumn{6}{c|}{Syn. HFID$\downarrow$} & \multicolumn{5}{c}{Real HFID$\downarrow$}
    \\ \cline{2-7} \cline{8-12}
    \makebox[0.04\textwidth][l]{Methods}  & 
    \makebox[0.05\textwidth][r]{Total}   & \makebox[0.04\textwidth][r]{0-10\%}  &
    \makebox[0.04\textwidth][r]{10-20\%} & \makebox[0.04\textwidth][r]{20-30\%} & 
    \makebox[0.04\textwidth][r]{30-40\%} & \makebox[0.04\textwidth][r]{40-50\%} &
    \makebox[0.05\textwidth][r]{Total}   & \makebox[0.04\textwidth][r]{0-10\%}  & 
    \makebox[0.04\textwidth][r]{10-20\%} & \makebox[0.04\textwidth][r]{20-30\%} & 
    \makebox[0.04\textwidth][r]{30-40\%} 
    \\ \midrule
    PConv       & 20.56   & 11.47   & 28.13    & 46.87    & 61.58    & 118.0   
                & 44.43   & 34.49   & 53.10    & 70.80    & 121.6    
                \\
    MAT         & 19.52   & 11.81   & 27.80    & 45.36    &  \underline{57.08}    & \underline{102.7}  
                & 47.85   & 39.40   & 56.02    & 73.56    & 123.9    
                \\ 
    TFill       & \underline{19.08}   & \underline{10.17}    & \underline{24.69}    & \underline{43.33}    & 59.30   & 108.55 
                & \underline{41.59}   & \underline{32.16}    & \underline{48.41}    & \underline{69.25} & \underline{114.6}   
                \\ \midrule
    DMAT(base)  & 17.51   & 9.91     & 24.01    & 40.17    & 55.02    & \textbf{92.56} 
                & 42.62   & \textbf{31.94}    & 49.73    & 63.05    & \textbf{109.21} 
                \\
    DMAT        & \textbf{16.34}   & \textbf{8.94}    & \textbf{22.34}    & \textbf{37.65}    & \textbf{51.14}    & 93.65  
                & \textbf{39.43}    & 30.16    & \textbf{45.43}    & \textbf{59.37}     & 109.43
                \\ \bottomrule
        \end{tabular}
        \caption{
        Performance comparisons on the AHP dataset with ground-truth amodal and modal masks in terms of HFID. '$0-10\%$' represents the occlusion rate while `total' denotes all images.
        }
        \label{tab:HFID_AHP}
    \end{center}
\end{table*}

In this supplemental material, we provide more performance comparison and visualization results.

First, we give the detailed human FID (HFID) performance variation with respect to the occlusion rate in Table \ref{tab:HFID_AHP}. The HFID only measures the similarity of human regions between the ground-truth images and the de-occluded results. As illustrated in this Table, our DMAT achieves the best performance for nearly all rates. Sometimes, the baseline obtains a little better results, but the gap is very small. In total, the proposed DMAT generates the smallest HFID and the gain over other methods are larger than 2\%. 

Next, the training computational complexity (FLOPs) and parameters of a model are important since larger model with more training parameters tends to produce better results. However, more computation resources are needed. To illustrate this, we further compare the training FLOPs and parameters of different methods. Note that the SAM is only used to generate the occlusion mask, which can be obtained once before model training and fixed during training. Therefore, we don't count the FLOPs and parameters of the SAM part in our model. Specifically, we use the command line tool "fvcore" to measure the model's \linebreak\newpage\noindent forward FLOPs and parameters (Params), where the mini-batch size is set to 1. The results are shown in Table \ref{tab:FLOP}. In this 
Table, we can see that the DMAT achieves the best results with relatively lower FLOPs and Params, which validate the efficiency of the proposed modules. 

\begin{figure*}[t]
    \centering
    \includegraphics[scale=0.5]{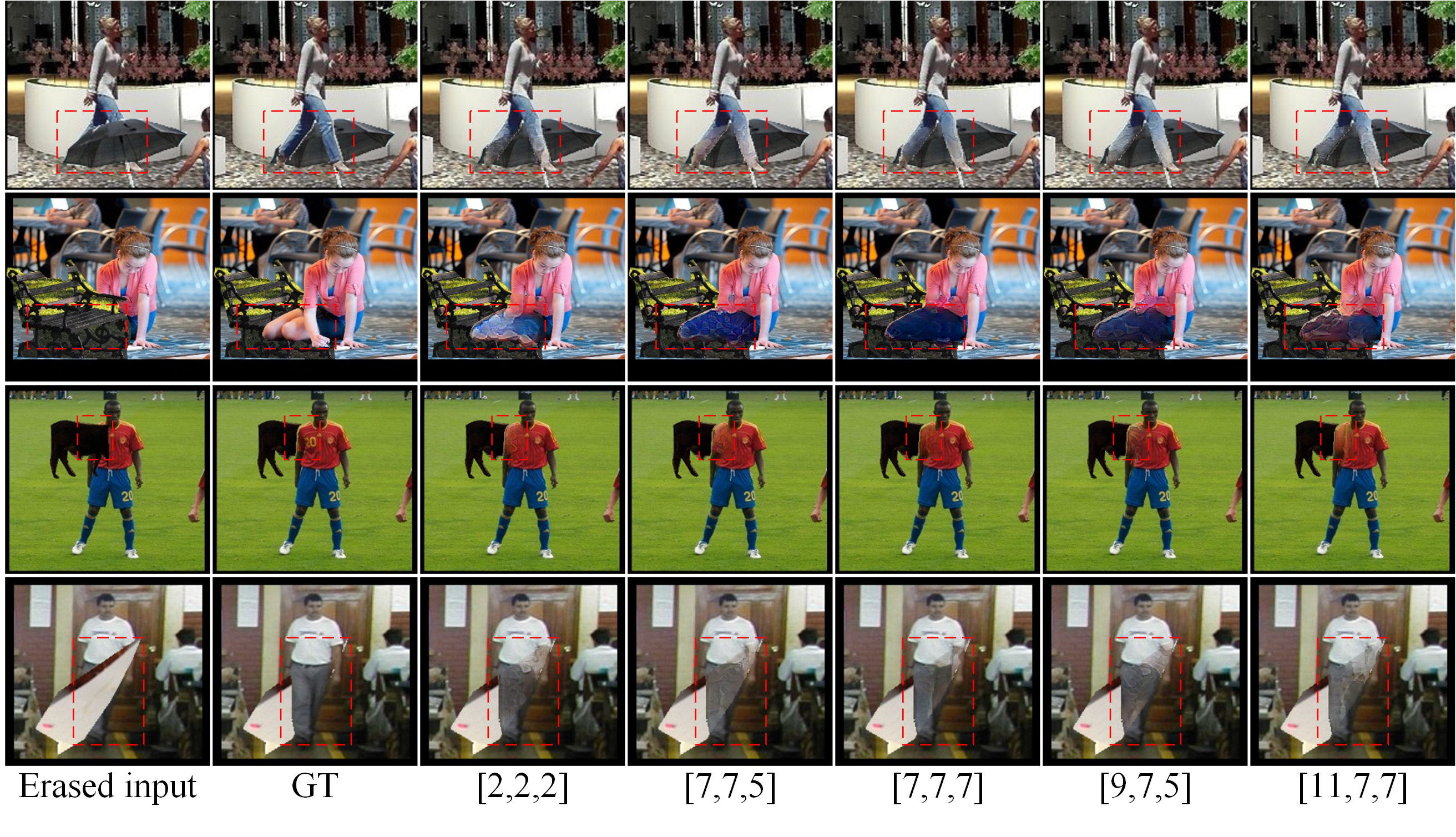}
    \captionof{figure}{Samples visualization using different convolution kernel sizes in ECH.}
    \label{fig:ECH_instance}
\end{figure*}

Moreover, in Figure \ref{fig:ECH_instance}, we show some generated samples with different kernel sizes in ECH. As seen from this figure, different strategies for the kernel sizes lead to various results. In detail, although some compared strategies produce plausible results, the de-occluded parts contain obvious checkerboard artefacts or wrong color, like the color of trousers in the first sample. In contrast, using [7, 7, 7] for the kernel size obtains the best performance, which addresses the checkerboard artefacts. This shows the effectiveness of the proposed ECH, which is similar to the FID comparison in Table 3 of our paper.

\begin{figure*}[t]
    \centering
    \includegraphics[width=0.91\textwidth]{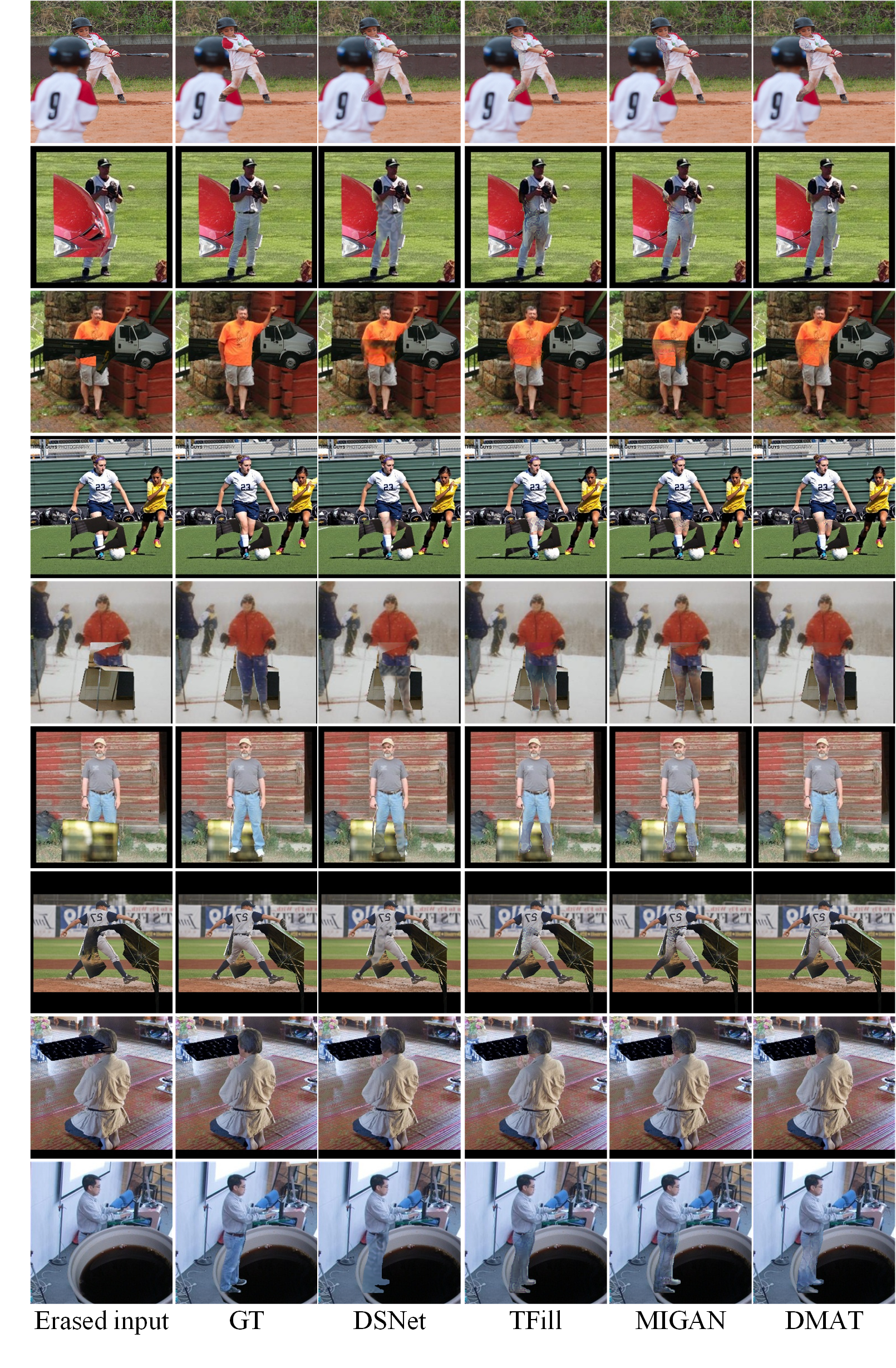}
    \caption{More visualization examples.}
    \label{fig:more_instance}
\end{figure*}

Finally, we provide more visualization samples in Figure \ref{fig:more_instance}. In these samples, the proposed DMAT gets the best appearance recovery effect. In contrast, the results by compared methods present checkerboard artefacts and inconsistent color, which may lead to mistakes.

The code will be available after acceptance.

\begin{table*}[h]
\begin{center}
    \begin{tabular}{c|cc|cc}
    \toprule 
    \multirow{2}{*}{} & \multicolumn{2}{c|}{FID $\downarrow$} & \multirow{2}{*}{\makecell[c]{FLOPs \\ (G)}} & \multirow{2}{*}{\makecell[c]{Params \\ (M)}} \\ \cline{2-3}
    \makebox[0.2\textwidth][c]{Methods} & \makebox[0.15\textwidth][c]{Syn.} & 
    \makebox[0.15\textwidth][c]{Real} &
    \makebox[0.15\textwidth][r]{} & \makebox[0.15\textwidth][r]{} \\
    \midrule
    PConv      & 15.34 & 19.74 & 19.18 & 51.56 \\
    DeepFillv2 & 16.60 & 15.92 & 162.2 & \textbf{3.04}  \\
    DSNet      & 13.23 & 14.02 & 24.90 & 74.14 \\
    MAT        & 15.08 & 16.72 & 140.2 & 56.66 \\
    TFill      & \underline{12.93} & \underline{11.65} & 153.4 & 12.11 \\
    FcF        & 17.42 & 12.55 & 40.58 & 71.92 \\
    MIGAN      & 16.03 & 11.73 & \textbf{11.35} & 52.71 \\ \midrule
    DMAT(base) & 11.83 & 11.43 & \underline{11.75} & \underline{11.05} \\
    DMAT       & \textbf{10.99} & \textbf{8.75}  & 14.04 & 12.02 \\
    \bottomrule
    \end{tabular}
    \caption{HFID performance comparisons on the AHP dataset. 'Syn.' and 'Real' denote the synthesized and real validation set respectively. Note that the FLOPs and Params are computed during model training.}
    \label{tab:FLOP}
\end{center}
\end{table*}

